\documentclass[review]{elsarticle}
\usepackage{lineno,hyperref,graphicx,float}
\usepackage{multirow}
\usepackage{array}
\usepackage{subfig}
\usepackage{booktabs}
\usepackage[fleqn]{amsmath}
\modulolinenumbers[5]
\journal{Journal of \LaTeX\ Templates}
\begin{document}

\begin{frontmatter}

\title{Real-world plant species identification based on deep convolutional neural networks and visual attention}
\author[a]{Qingguo Xiao\corref{mycorrespondingauthor}}
\cortext[mycorrespondingauthor]{Corresponding author.}
\ead{qgxiao@tongji.edu.cn}

\author[a]{Guangyao Li} % mymainaddress
\author[a]{Li Xie}
\author[a]{Qiaochuan Chen}
%\author[b]{Mang Xiao}
%\author[c]{Li Yi}

\address[a]{College of Electronics and Information Engineering, Tongji University, Shanghai, China} %
%\address[b]{School of Computer Science and Information Engineering, Shanghai Institute of Technology, Shanghai, China}
%\address[c]{Intel Asia-Pacific Research and Development Ltd.}

\begin{abstract}
This paper investigates the issue of real-world identification to fulfill better species protection. We focus on plant species identification as it is a classic and hot issue. In tradition plant species identification the samples are scanned specimen and the background is simple. However, real-world species recognition is more challenging. We first systematically investigate what is realistic species recognition and the difference from tradition plant species recognition. To deal with the challenging task, an interdisciplinary collaboration is presented based on the latest advances in computer science and technology. We propose a novel framework and an effective data augmentation method for deep learning in this paper. We first crop the image in terms with visual attention before general recognition. Besides, we apply it as a data augmentation method. We call the novel data augmentation approach attention cropping (AC). Deep convolutional neural networks are trained to predict species from a large amount of data. Extensive experiments on traditional dataset and specific dataset for real-world recognition are conducted to evaluate the performance of our approach. Experiments first demonstrate that our approach achieves state-of-the-art results on different types of datasets. Besides, we also evaluate the performance of data augmentation method AC. Results show that AC provides superior performance. Compared with the precision of methods without AC, the results with AC achieve substantial improvement.
\end{abstract}  %    interdisciplinary collaborations in this challenging research field.    which is an important approach for deep learning   and valuable

\begin{keyword}
%\texttt convolutional neural network\sep ensemble learning\sep filter shape\sep image representation   % hai mei   gai
Plant protection, Image classification, Real-world recognition, Deep learning, Visual attention, Data augmentation   %  
\end{keyword}

\end{frontmatter}

\linenumbers

\section{Introduction}
Plants play an irreplaceable role in our world and they have direct effect in many domains such as agriculture, climate, ecological system and so on. Besides, they are the main source of food for human survival and development. Many problems such as habitat degradation, global warming, ecosystems destruction, environment worsen, species extinction, and so on  have something to do with plant protection. Plant species identification is the prerequisite for protection. There have been many research related to the issue. Method based on image classification is now considered to help improve the plant taxonomy. It is one of the most promising solutions among those related research work, as discussed in\cite{goeau2016plant}. And it has been a long term hot research issue.

Considering flowers and fruits of plants are seasonal, some researchers believe that leaves are more suitable for identification. In the early time, leaves are frequently used for computer-aided plant species classification. Most image-based identification methods and evaluation data proposed were based on leaf images\cite{kumar2012leafsnap,backes2009plant,cerutti2011parametric}. However, most leaf images are specimen or scanned at that time. The way to acquire samples is also strict. Afterwards, flowers begin to be employed\cite{nilsback2008automated,angelova2012development,nilsback2006visual}.

Obviously, approaches only with leaves or flowers are insufficient considering realistic plant identification and protection. More diverse parts of plants have to be considered for accurate identification, especially because it is not possible for plants to see their leaves all over the year. Compared with the photos token by realistic ways, the background in those datasets where the camera are closely to targets when people take pictures is simple. We believe that those are not real-world identification task. In this paper, we focus on  plant species identification especially realistic recognition. For a real-world plant identification task, plant image samples should include many parts such as fruits, branches, entire, apart from leaves and flowers. At the same time, the way to create and acquire plant images should not be strict. They can be snapshotted by different users and at different time and users can be at their own will. Image samples can be with complicated background. Besides, the scene includes not only indoor but also field. We believe that these indeed belong to real-world plant identification. The task of the identification is more challenging compared with tradition species identification while it is more valuable at the same time. And only real-world species recognition can realize better, convenient,and comprehensive plant protection.
%and only   valuable, meaningful  help  fulfill plant protection   which is more challenging and meaningful

In the last few years, a number of projects and organizations such as iNaturalist, Botanica can generate large amounts of biodiversity data.  Big biodiversity data can be available easier compared with the past\cite{agarwal2006first}. As it is convenient for a vast majority of people to snapshot plant images with mobile phones. The way to create and acquire plant images becomes easier and close to the condition of a real-world scenario. Besides, people like to share them and chat with each other on their personal social networks. %

\begin{figure}  %[!htbp]
\begin{center}
    \fbox{\includegraphics[width=0.6\linewidth]{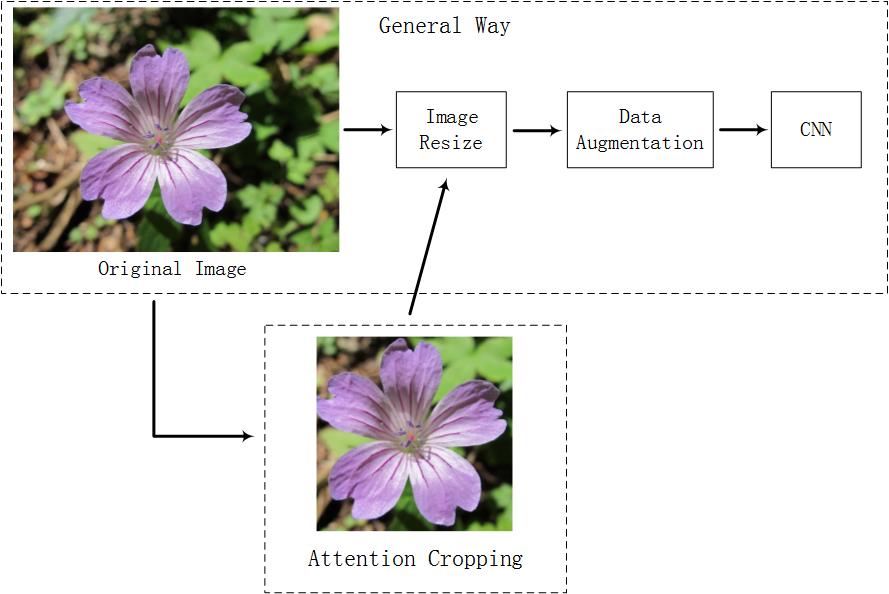}}
\end{center}
   \caption{Illustration of image classification in general way and our proposed method.}
\label{Fig 1}
\end{figure}
In this paper, we first review the related work about plant species recognition issue. We discuss commonly used approaches using leaf and flowers for tradition species identification. We then propose a novel framework and an effective data augmentation method to address the task of realistic recognition. As deep convolutional neural networks (CNNs) provide us an useful tool for large-scale image classification, we conduct our work basing on deep learning. As stated in \cite{Yarbus1967eye,neisser1967cognitive} , the salient objects which are to be recognized in an image are focused on in terms of our human visual attention . We crop the image in terms of our visual attention and name the operation attention cropping (AC). AC is accomplished with the generated saliency map using saliency detection approach. As we all known, data augmentation is an important operation in deep learning. Here, we apply AC as a data augmentation method for deep learning. The schematic diagram of our framework is shown in Figure ~\ref{Fig 1}.  We validate our proposed approach through a series of comparisons and results show that superior results are achieved. %significantly.

\section{Related work}
\begin{figure}%[!htbp]
\begin{center}
    \fbox{\includegraphics[width=0.8\linewidth,height=6cm]{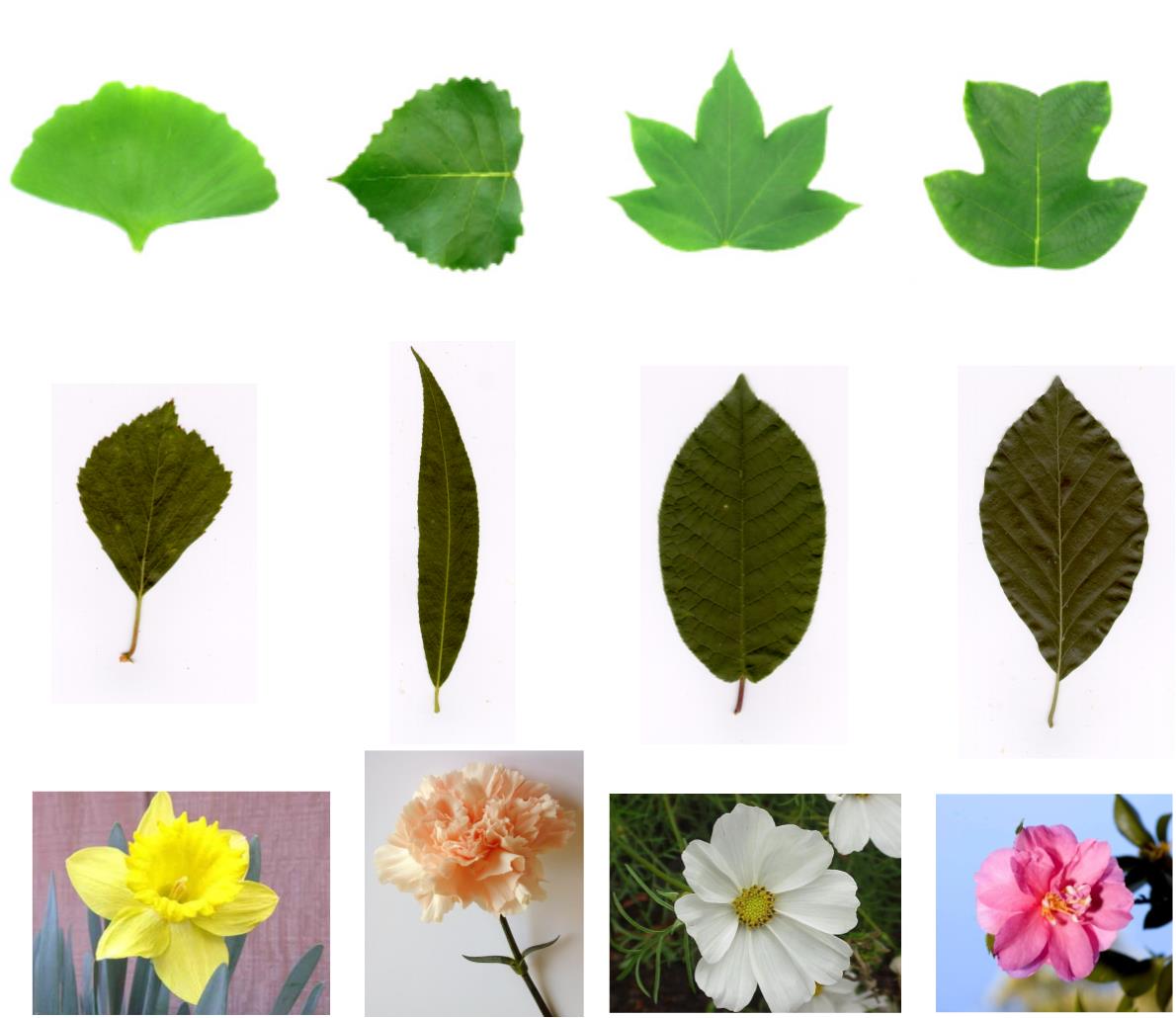}}
\end{center}
   \caption{Samples of Flavia leaf, Swedish leaf, and Oxford flower. The top are from Flavia leaf, the samples in second row are from Swedish leaf. Correspondingly, the last row are from Oxford flower.}
\label{Fig 2}
\end{figure}
There are amounts of plant identification approaches that use digital images. As stated above, early algorithms are mainly with leaves. Flavia\cite{Flavia} and Swedish leaf database\cite{Swedish} are two typical leaf datasets. Samples of Flavia and Swedish leaf are shown in Figure ~\ref{Fig 2}. \cite{wu2007leaf} employed Probabilistic Neural Network(PNN) with image and data processing techniques to implement a general purpose automated leaf recognition for plant classification. Using Artificial Neural Network(ANN), \cite{soderkvist2001computer} designed a computer vision classifier to identify the different Swedish tree classes in terms of their leaves. Many methods employed shape or curvature features as plants are basically classified according to the shapes of their leaves\cite{bai2010learning,neto2006plant,du2006computer,wang2014hierarchical,ling2007shape}. Shape or curvature features are relatively discriminative for leaf images  according to the theory of plant shape taxonomy. It is efficient especially when image contents are extreme simple. Two-dimensional multifractal detrended fluctuation analysis is used for plant classification in \cite{wang2015two}.

Afterwards, flower image samples begin to be employed. Oxford flower is applied in many related researches\cite{nilsback2006visual,nilsback2007delving,nilsback2008automated,nilsback2010delving}. \cite{nilsback2006visual} developped a visual vocabulary that explicitly represents the various aspects(colour, shape, and texture) that distinguish one flower from another. \cite{nilsback2008automated} investigated to what extent combinations of features can improve classification performance on a large dataset of flower classes. Samples of Oxford flower dataset are shown in the last row of Figure ~\ref{Fig 2}. The work of \cite{nilsback2007delving} and \cite{nilsback2010delving} is similar. They focused on algorithms for automatically segmenting flowers in colour photographs.

\begin{figure}%[!htbp]
\begin{center}
    \fbox{\includegraphics[width=0.8\linewidth]{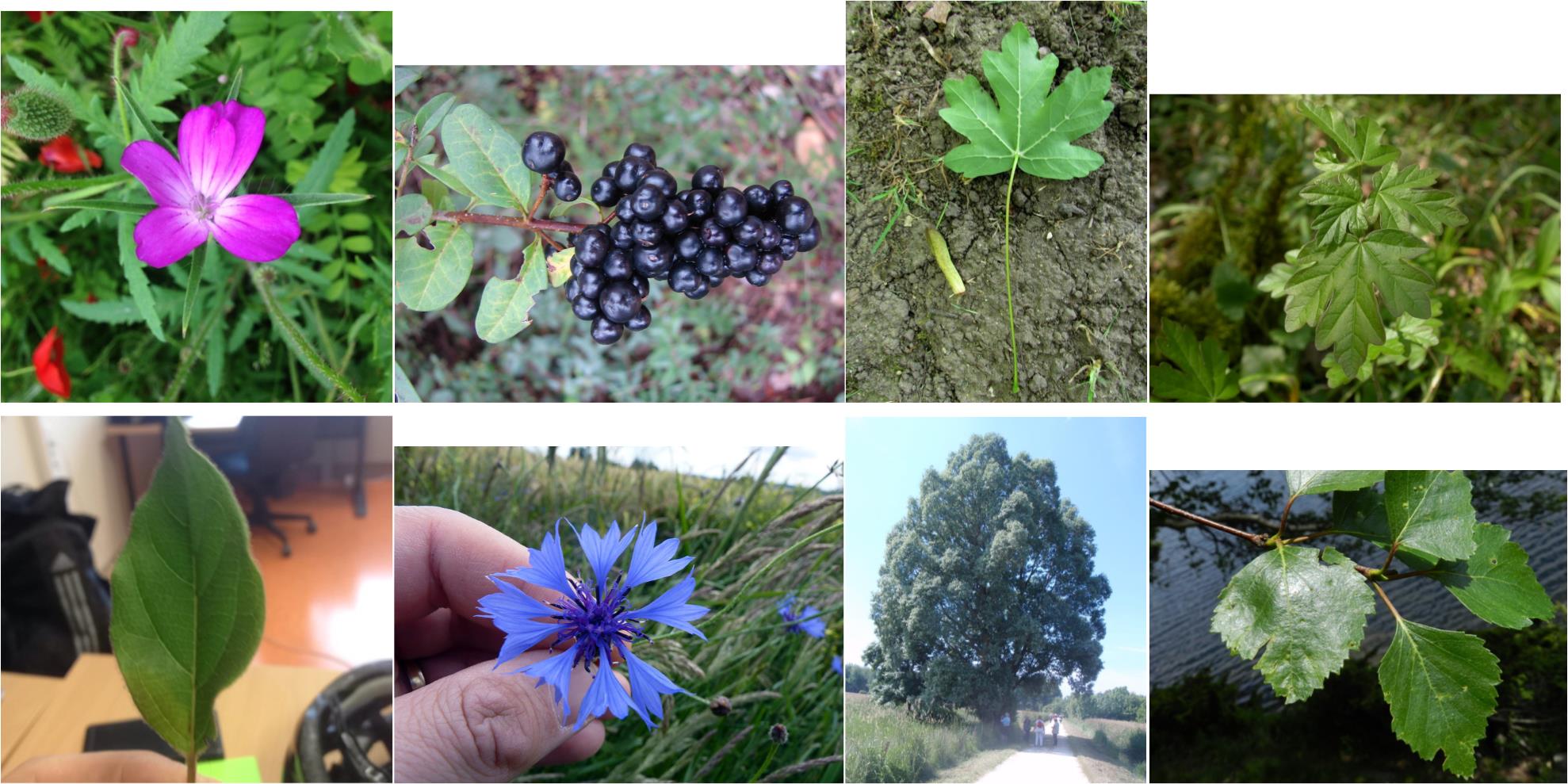}}
\end{center}
   \caption{Samples of PlantCLEF. There are different view types and the samples are close to realistic scenarios.}%including leaf, flower, fruit, and entire are shown Samples of
\label{Fig 3}
\end{figure}
As we stated above, more parts of a plant such as flowers, leaves, fruits, branches, stem, should be used for realistic plant species recognition. The image data can be collected with a number of different contributors, cameras, areas, periods of the year, individual plants, etc. Recently, an image-based plant identification dataset called PlantCLEF was initially conducted. It is near to real-world conditions. Our work is with PlantCLEF in this paper. Samples of the dataset are shown in Figure ~\ref{Fig 3}. The schematic diagram of our proposed method is shown in Figure ~\ref{Fig 1}. We will describe our approach and work in detail in the next.

\section{Approach}

\subsection{Attention cropping}

\begin{figure}[!htbp]
\begin{center}
    \subfloat{\includegraphics[width=0.32\linewidth]{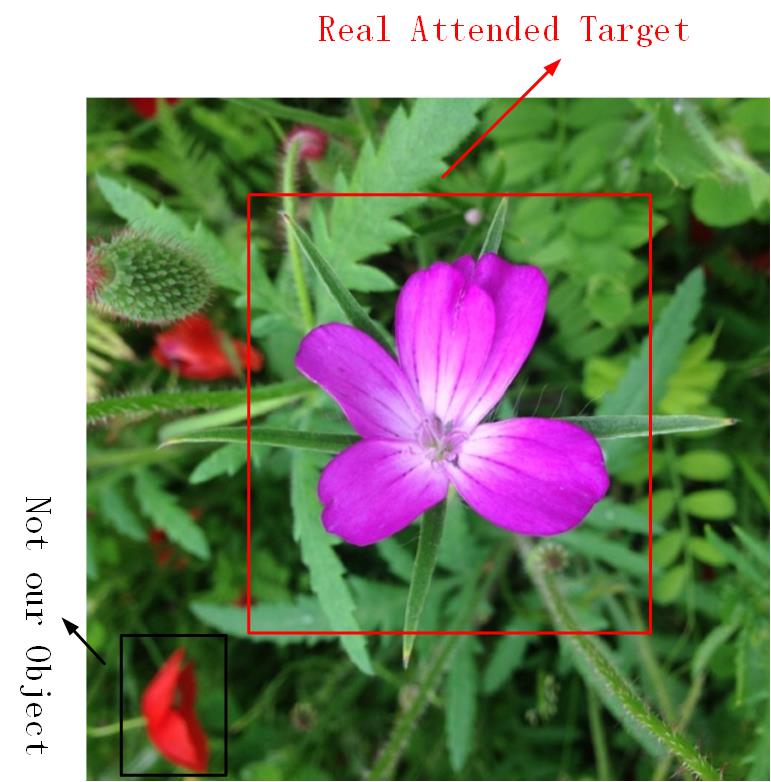}}\
    %\hspace{1in}
    \subfloat{\includegraphics[width=0.32\linewidth]{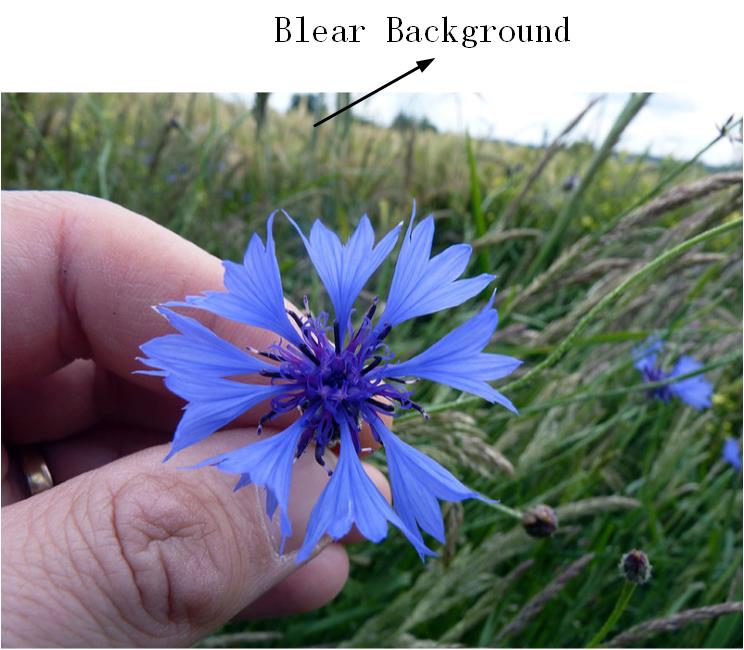}}\
    \subfloat{\includegraphics[width=0.32\linewidth]{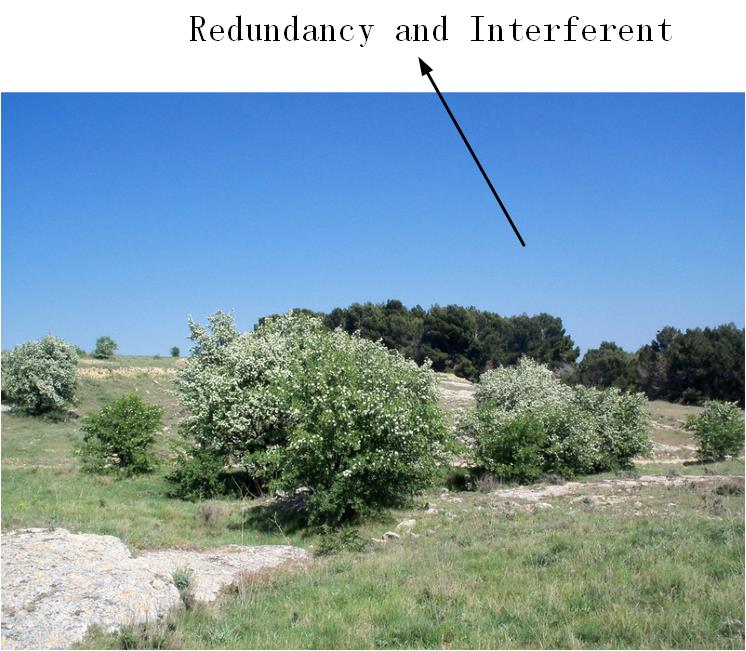}}
\end{center}
   \caption{Illustration of why we use attention cropping. (a) Obviously, the center body indicated with a red box is to be recognized and the left bottom tiny flower marked with a black rectangle is not the real target. (b) The background is blear and should be suppressed. (c) The blue sky is meaningless background. The behind trees in the distance are interferents as they are similar to the targets. }
\label{Fig 4}
%\label{fig:onecol}
\end{figure}

The background of images taken in real-world ways is usually complicated. A real-world plant image contains more than one object, i.e. target plants and background objects(small stones, ruderals, branches, non-target leaves and other interferents). Moreover, target plants are possibly touching or covering the background objects. However, the salient objects what we pay attention to are to be recognized in an image. For an image, visual attention facilitates our ability to rapidly locate the most important information in a scene\cite{Yarbus1967eye} and  the most useful point are focused on with our attention at first sight for an given object\cite{neisser1967cognitive}. As demonstrated in Figure ~\ref{Fig 4} (a), the centered object indicated with a red box is our real interesting target and to be recognized. The left bottom one boxed with a black rectangle is not an object for recognition  and it should be neglected. In addition to the interference of the non-target, there are also interferents and non-valuable redundance as demonstrated in Figure ~\ref{Fig 4} (b) and (c). The object is recognized  only with the sketchy and concentrated screenage or info borne in our mind although there are lots of contents. Other non-salient parts are neglected or ignored. We even do not have any aware of the redundance during the first judgement.  %Basing on this assumption, we believe that
% http://blog.csdn.net/daxiamit/article/details/7532915   Here, we base on an assumption  Considering our human recognition tactics,

Here, salient regions where we attend are got with saliency detection ways. Using the approach described in \cite{li2013visual}, we obtained the image saliency map. \cite{li2013visual} shows that the convolution of an image amplitude spectrum with a low-pass Gaussian kernel is equivalent to an image saliency detector.

\begin{equation}
     S=\mathcal{F}^{-1}\left\{{{A_S}(u,v)e^{i\cdot{{P}(u,v)}}}\right\}
\end{equation}
where ${P}$ is the original phase spectrum, ${A_S}$ is the resulting smoothed amplitude. ${A_S}$ is as follows:
\begin{equation}
   {A_S}(u,v)=|\mathcal{F}\{{f(x,y)}\}|\star{h}
\end{equation}

The saliency map is obtained by reconstructing the 2D signal using the original phase and the amplitude spectrum. The low-pass Gaussian kernel scale $h$ is filtered at a scale selected by minimizing saliency map entropy. After the proper scale is specific, the resulting smoothed amplitude ${A_S}$ will be computed according to the formula (2). And Hypercomplex Fourier Transform (HFT) is employed to replace standard Fourier Transform (FT) to performs the analysis in the frequency domain.
After phase spectrum ${P}$ and the resulting smoothed amplitude ${A_S}$ are computed, then the saliency map is got in terms of the formula (1). The low-pass Gaussian kernel scale $h$ can be also set by hand. To get a better saliency map, it is got based on spectrum scale-space analysis to find a proper scale. Then the got saliency map will be used for segmentation.

After that, image segmentation is carried out for generating the regions of interest (ROI) for recognition. K-means is used to perform the segmentation. Given a set of observations ${(\textbf{x}_1, \textbf{x}_2,...,\textbf{x}_n)}$ , where each observation is a $d$-dimensional real vector, k-means clustering aims to partition the $n$ observations into $k (k \leq n)$ sets $\textbf{S} = {S_1, S_2,..., S_k}$ so as to minimize the within-cluster sum of squares (WCSS) (i.e. variance). Formally, the objective is to find:
\begin{equation}
   \mathop{\arg\min}_{S}\sum^k_{i=1}\sum_{\textbf{x}\in{S_i}}\lVert{\textbf{x}-\textbf{u}_i}\rVert^2
\end{equation}
where $\textbf{u}_i$ is the mean of points in $S_i$. The operation of cropping is then implemented in terms of the segmentation results. It is defined as:
%Non-salient parts such as distant background, indistinct surroundings, and corners are tailored out finally. The cropping operation is defined as: 

\begin{equation}
   \textbf{\emph{b}}_{s}=argmin\ {f(\textbf{\emph{I}}_{seg}>\emph{th})} , \qquad
   \textbf{\emph{b}}_{e}=argmax\ {f(\textbf{\emph{I}}_{seg}>\emph{th})}
\end{equation}
\begin{figure*}%[ht]
\begin{center}
    \includegraphics[width=1.0\linewidth]{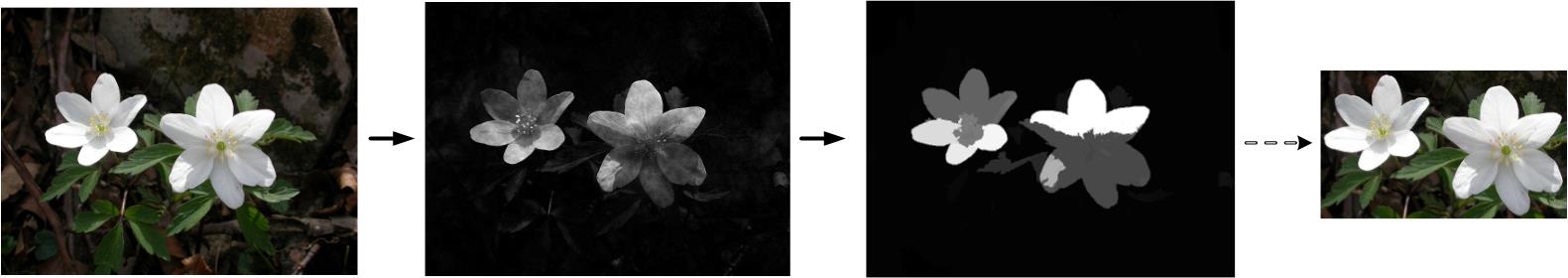}
\end{center}
   \caption{Flow chart of attention cropping of a flower (Anemone nemorosa L) sample. The first picture is the original image. The two intermediate pictures represent saliency detection and segmentation results respectively. The dotted arrow means cropping operation here. We crop the original image using the generated coordinates.}
\label{Fig 5}
%\label{fig:onecol}
\end{figure*}
%\begin{equation}
%   \textbf{\emph{b}}_{e}=argmax\ {f(\textbf{\emph{I}}_{seg}>\emph{th})}
%\end{equation}
where $\textbf{\emph{I}}_{seg}$ is the segmentation result vector, \emph{th} is a threshold which can control the degree of cropping, $\textbf{\emph{b}}_{s}$ is the start position of the target area and $\textbf{\emph{b}}_{e}$ is the end position. \emph{th} is got by multiplying the cluster number $N$ by parameter $\lambda$, which is the ratio of the clusters what we want to crop out to the total clusters.
%, \emph{th} can be computed as:
%\begin{equation}
%  \emph{th}=\lambda \times N
%\end{equation}  %  for example. the the cluster number $N$

\begin{figure}%[ht]
\begin{center}
\fbox{\subfloat[]{\includegraphics[width=1.0\linewidth]{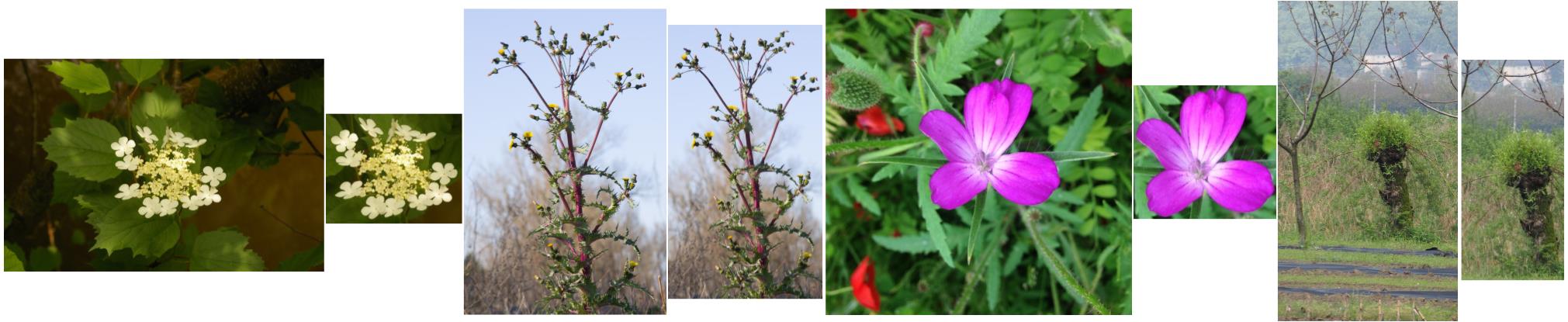}}}\\
\fbox{\subfloat[]{\includegraphics[width=1.0\linewidth]{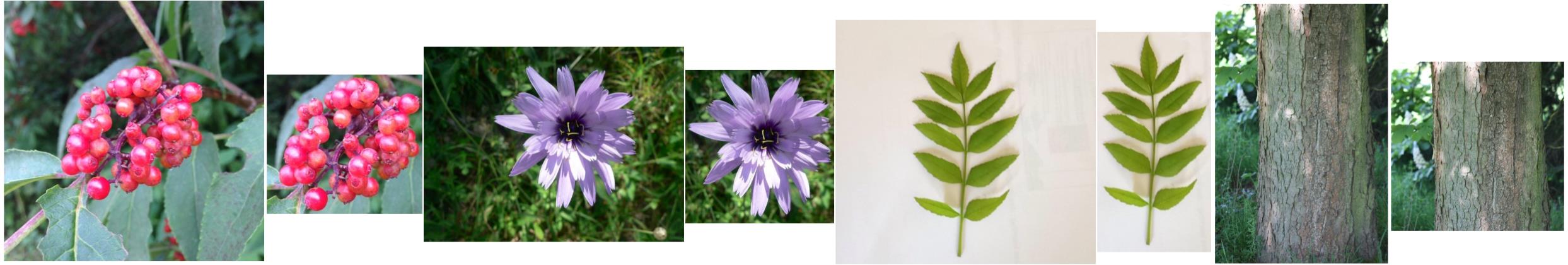}}}
\end{center}
   \caption{Comparisons between original images and final attention cropping results. The left is the original image and the right is the cropped one in every comparison.}
\label{Fig 6}
%\label{fig:onecol}
\end{figure}

The cluster number $N$ and parameter $\lambda$ are set empirically in this paper. For other methods, these can be adaptive values. We get the corresponding coordinates of the ROI in accordance with the above computed results $\textbf{\emph{b}}_{s}$ and $\textbf{\emph{b}}_{e}$ . % by hand
In the end, the original image is cropped in terms of the generated corresponding coordinates to obtain the attended image regions. Non-salient parts such as distant background, indistinct surroundings, and corners are tailored out finally. Take a sample of flower (Anemone nemorosa L) for example, we illustrate attention cropping in Figure ~\ref{Fig 5}. The intermediate results are also shown. In addition, comparisons between original images and final attention cropping results are shown in Figure ~\ref{Fig 6}.
%AC helps us focus on the objects what we are most interested for a given image especially for a realistic image.
% We only demonstrate More results are shown in Figure ~\ref{Fig 8} in Section 4.   %This is shown in Figure ~\ref{Fig 6}.

%there are three steps totally .
% image segmentation is our  after saliency detection and the cropping is the final step of the algorithm and it is implemented in terms of the segmentation results. Similar as random cropping, its goal is to fulfill data augmentation for deep learning to train convolutional neural networks. % 
%Then new samples are created for training convolutional neural networks.
%and these samples
% to create new samples for training convolutional neural networks
%  the final goal is to fulfill  AC and  get the real target .  for deep learning. next   data augmentation. horizontalflip, randomcrop
% the final   % include the step details , 

\subsection{Convolutional Neural Network}

Convolutional neural networks (CNNs) demonstrate impressive results in image classification. In this paper, the approach of CNN is adopted to facilitate the classification. CNNs directly use raw image as an input and image category as an output so that forms an end-to-end system. They learn image features from training of network. A CNN network consists of three types of layers, including convolutional layers, pooling layers, and at least one final fully connected layer. The outputs are generally normalized with a Softmax activation function and therefore approximate posterior class probabilities. For a given output feature map
$x_{ij}$, the activation of Softmax is as follows:
\begin{equation}
p_{ij}=\frac{exp(x_{ij})}{\sum_{k}{exp(x_{kj})}}
\end{equation}
$p_{ij}$ denote the probability of map $i$ mapping to class $j$ and subject to the constraints that $\sum_{i}{p_{ij}}=1$ and $0\leq p_{ij}\leq{1}$.

\section{Experiments and analysis}

We first evaluate the performance of our method on PlantCLEF. The employed dataset is composed of about one hundred thousand pictures belonging to 1000 species. Each picture belongs to one and only one of the seven types of views reported in the meta-data. These types are entire plant, fruit, leaf, flower, stem, branch, leaf scan.

An originality of PlantCLEF is that its "social nature" makes it closer to the conditions of a real-world identification scenario: (i) images of the same species come from distinct plants living in distinct areas, (ii) pictures are taken by different users that might not use the same protocol of image acquisition, (iii) pictures are taken at different periods in the year.

%It is worth mentioning that the latest version of the dataset is not for pure classification task. In addition to pure classification, it is for addressing the issuse of recognizing unknown and never seen categories as well. We do not adopt the latest dataset in this paper. Here, we merely focus on classification task and use PlantCLEF 2015, which is similar to ImageNet dataset.
%%\noalign{\smallskip}

\subsection{Experiments setup}
We then perform our proposed novel framework and data augmentation method of attention cropping (AC). The experiments are conducted with two NVDIA Geforce GTX-1080 GPUs. The deep learning software tool is Pytorch. MATLAB is also used in our experiments. The preprocessing which is attention cropping is fulfilled with MATLAB. For simplicity, we set the total cluster number $N=3$ and cropping ratio $\lambda = 1/3$.
%Figure ~\ref{Fig 8}  gives some results of attention cropping. There are several different view types and scenarios as shown. This is popularly used to train networks in the vision community.

Two neural networks of ResNet50 and Inception v3 are chosen to conduct the experiments to evaluate the performance of the proposed framework. ResNet50 is a kind of residual network. Residual networks are similar as VGG but with learning residual functions with reference to the layer inputs to ease the training of deeper neural networks. So they are not sequential models. ResNet50 is with a depth of 50 layers. Its major kernel style is 3$\times$3. Inception v3 introduces inception modules. Inception modules help increase the width of the network. It is a convolutional block which is constituted by different kinds of convolutional kernels. Apart from  3$\times$3 kernel is employed which is common used, other types such as 1$\times$7,7$\times$7,1$\times$1,1$\times$3 and so on are also adopted for constructing networks. Big and small convolutional kernels are used together in one block.  Big convolutional styles and feature maps are with little number of kernels and small convolutional styles and feature maps are with large number of kernels.

Besides, experiments without AC are also  carried out to prove the effectiveness of the proposed data augmentation method.  Considering the training efficiency, the image preprocessing is carried out off-line. So, there are two different kinds of training datasets totally. One is the original and the other is got by using AC. Then the training is carried out with normal procedures. We set batch size 64 and use a linear decaying learning rate with a factor of 10 every 30 epochs. The initial learning rate is 0.1 and the total epochs are 90. Other parameters such as momentum and weight-decay are 0.9 and 1 $\times$ $10^{-4}$ respectively.

The input end of deep learning network generally requires input of the same size picture. All images will be resized to the given size after data augmentation such as flipping, random cropping which is popularly used to train networks in the vision community. A crop of random size of (0.08 to 1.0) of the original size and a random aspect ratio of 3/4 to 4/3 of the original aspect ratio is made in random cropping. Different from that, AC data augmentation is in terms of vision cognition. It is more essential. More important and interesting information is focused on.  And this crop is finally resized to different given size in terms of different deep learning networks. For resnet50, it is 224$\times$224$\times$3. And for inception v3 it is 299$\times$299$\times$3.

Then these all samples are shuffled for training. The number of the neurons of last fully connected (FC) layer are set N, which is the number of the categories. The loss function is cross-entropy loss. It is formulated as follows:
%  (we set 224 in our experiments)  Experiments are on .
\begin{equation}
     l=-\sum^K_{k=1}log({p(k)})q(k)
\end{equation}
where $p(k) \in [0,1]$ is the predicted probability of the input belonging to class $k$, $q(k)$ is the ground truth distribution. %Evaluation metric is mean average precision.
\begin{figure}[!htbp]
\begin{center}
    \subfloat[Leaf (Acer campestre L)]{\includegraphics[width=0.7\linewidth]{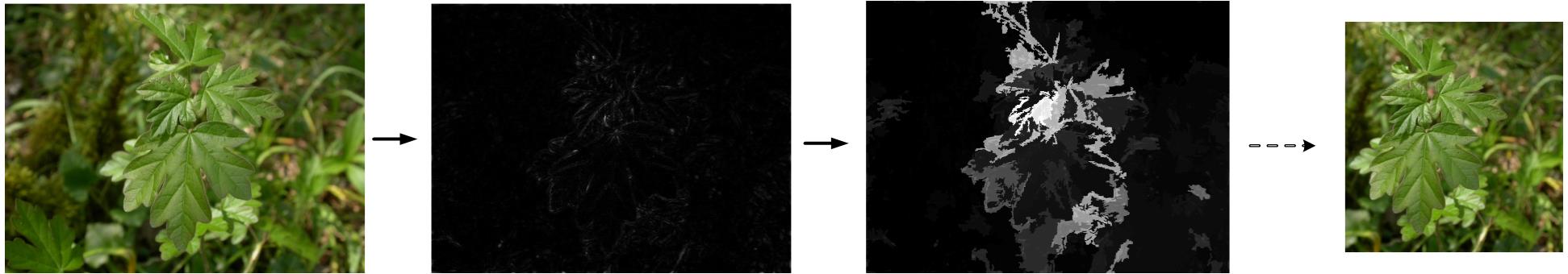}}\\
%    \caption{Example of caption. It is set in Roman so that mathematics}
    \subfloat[LeafScan (Acer pseudoplatanus L)]{\includegraphics[width=0.7\linewidth]{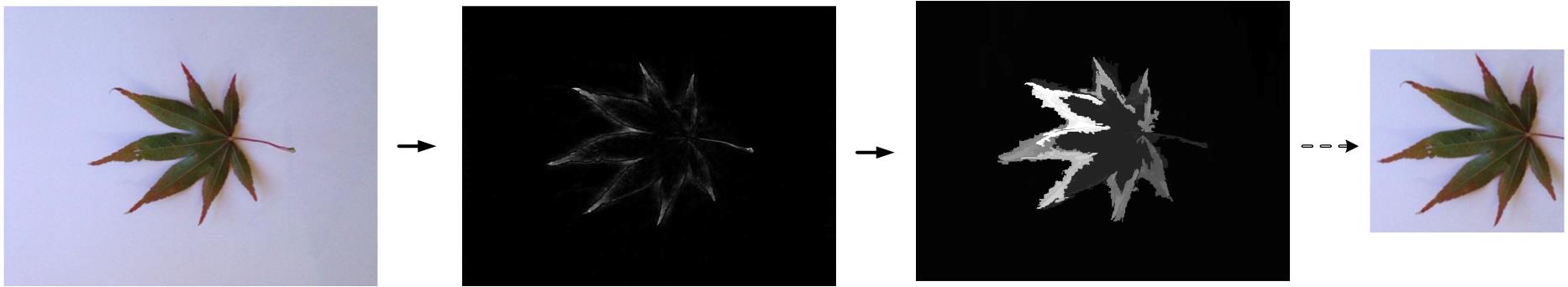}}\\
    \subfloat[Fruit (Malus domestica Borkh)]{\includegraphics[width=0.7\linewidth]{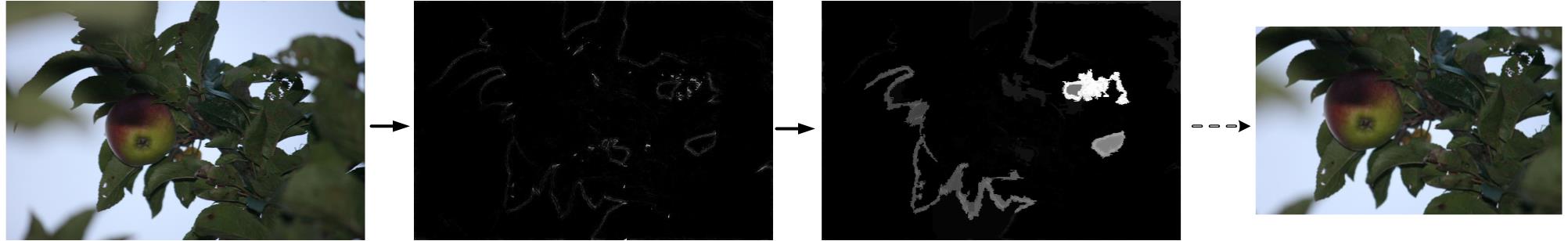}}\\
%    \subfloat[Entire (Carlina acanthifolia All)]{\includegraphics[width=0.8\linewidth]{entire.jpg}}\\
\end{center}
   \caption{Demonstration of attention cropping on different plant view types and species. The bracket notation is its specie name.}
\label{Fig 7}
%\label{fig:onecol}
\end{figure}

The learning method  is stochastic gradient descent (SGD). And it is kept same for all the two employed deep learning networks. Considering the problem of minimizing an objective function that has the form of a sum:
\begin{equation}
     Q(w)=\frac{1}{n}\sum_{i=1}^{n}{Q_i(w)}
\end{equation}
where the parameter $w$ which minimizes $Q(w)$ is to be estimated. $Q(w)$ is the loss or cost function. Each summand function $Q_{i}$ is typically associated with the $ i-th$ observation in the data set (used for training). Gradient descent method would perform the following iterations:
\begin{equation}
     w=w-\eta \nabla{Q(w)} = w- \eta \sum_{i=1}^{n}\nabla{Q_i(w)}/n
\end{equation}
where $\eta$  is a step size (sometimes called the learning rate in machine learning). In stochastic (or "on-line") gradient descent, the true gradient of $Q(w)$ is approximated by a gradient at a single example in each iteration and the sample is selected randomly (or shuffled) instead of as a single group (as in standard gradient descent) or in the order they appear in the training set.

% stochastic gradient descent

%and parameter settings of deep learning network are not given. ResNet50  % banmain de wenti ,women tigongle fujian, jian dan jie shixai zhge suanfa. jiage gongshi ye keyi. !
%are two popular networks.  % wei le fang bian kan

   %  . is a block. this add the width of networks.
%kernel, style and numbers  % repeat again, jiushi ba zijizhdiaode ,tamen buzhidaode ,zhuangxia bi,!

\subsection{Results and analysis}
Attention cropping (AC) is implemented after parameters are setup. Apart from flowers, AC also performs well on leaf, leafscan, fruits view types and \emph{et al}. The AC flow charts of more species and view types including leaf, fruit, leafscan, and entire are demonstrated as in Figure ~\ref{Fig 7}.  % realistic image fenlei fang lingyige limian haohaoaxie  such as leaf, leafscan, fruit, and \emph{et al}
%AC  performs well on flower, leaf, leafscan, and fruits view types.
\begin{figure}[!htbp]
\begin{center}
   % \fbox{\includegraphics[width=0.8\linewidth]{jgt.jpg}}
    \fbox{\subfloat[]{\includegraphics[width=1.0\linewidth]{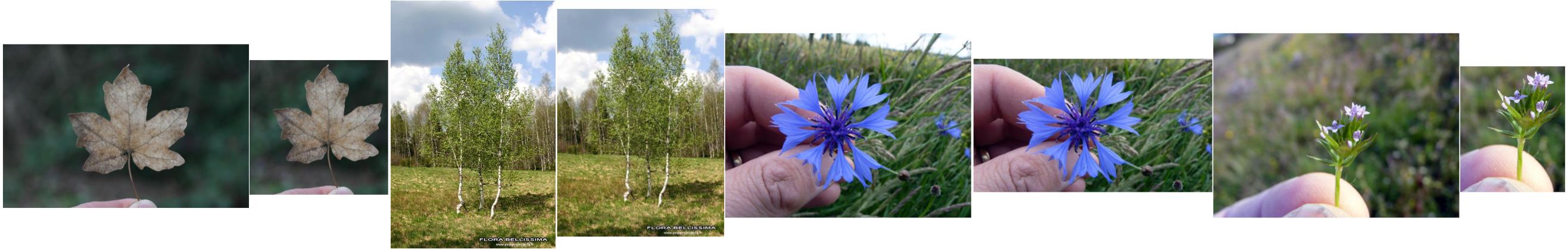}}}\\
    \fbox{\subfloat[]{\includegraphics[width=1.0\linewidth]{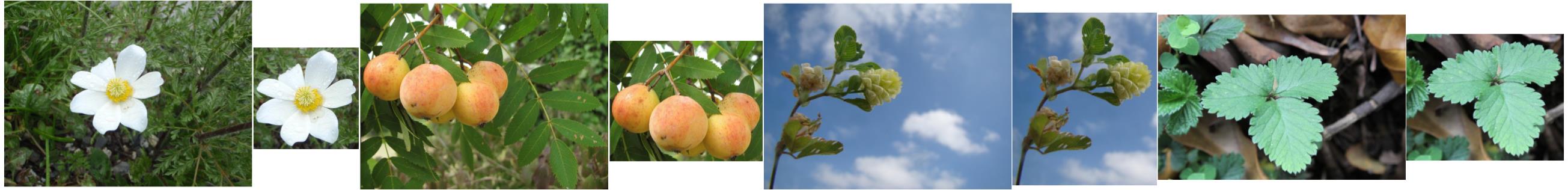}}}\\
  %  \fbox{\subfloat[]{\includegraphics[width=0.8\linewidth]{newjgt3.jpg}}}\\
    \fbox{\subfloat[]{\includegraphics[width=1.0\linewidth]{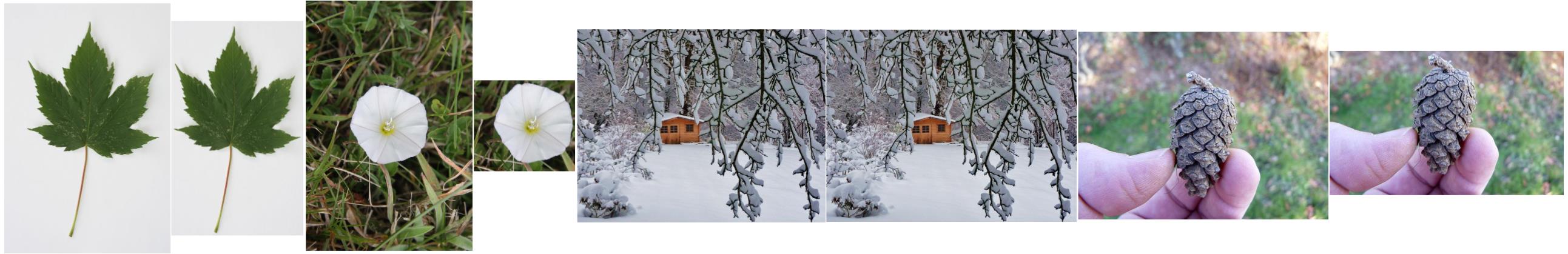}}}\\
\end{center}
   \caption{Comparisons between original images and final attention cropping results. Different view types and scenarios are included.}
\label{Fig 8}
\end{figure}
AC results of different plant species and view types are shown in Figure ~\ref{Fig 8}. There are several scenarios and they are different from each other. Results show that the targets and foregrounds are reserved and the distant surroundings and redundancies are tailored out after attention cropping. The most important information and what we are most interested are facilitated to be located using AC to fulfill real-world identification although there are different scenarios  and the background is complicated in realistic images.
%  keep with most interested  the targets or foregrounds are highlighted or the surroundings are distant, our method performs nicely. AC helps us facilitates our ability to rapidly locate
%we focus on    the most important information. As stated above, realistic image classification is more challenging.

%the flow chart
The final identification is performed using pre-trained deep convolutional neural networks. Accuracy of different methods on the test set of PlantCLEF are shown in Table 1. Attention cropping is abbreviated as "AC".   %In Table 1, we show comparison results of different methods on the test set on PlantCLEF.
%\documentclass[UTF8]{ctexart}   The results with attention cropping method are marked plus "AC" as shown.
%\usepackage{booktabs} %
%\begin{document}

\begin{table}[!htbp]
\footnotesize
\caption{Classification results on PlantCLEF}
\label{Table1}
\begin{center}
\begin{tabular}{lc}
\toprule  
Method& Accuracy\\
\midrule  
KDES\cite{le2015mica}&0.194\\
Reyes \emph{et al.}\cite{reyes2015fine}&0.486\\
CNN+FV+Bayesian\cite{champ2015comparative}&0.581\\
CNN+SVM\cite{ge2015content}& 0.590\\
Choi\cite{choi2015plant}&0.652\\
Ours(ResNet50 without AC)&0.636\\
Ours(Inception v3 without AC)&0.653\\
Ours(ResNet50 with AC)&0.680\\
Ours(Inception v3 with AC)&\textbf{0.695}\\
\bottomrule 
\end{tabular}
\end{center}
\end{table}

% The camera are closely to targets when people take pictures. And the background is relatively simple compared with the photos token by realistic way.  We can also benefit from training deeper architecture. The performance of ResNet50 and Inception v3 are better than that of GoogLeNet (Choi \emph{et al.}, also with ensemble approach), and AlexNet. It is noted that experiments are performed twice with the same platform and parameters to make fair comparisons considering the factors of different deep learning platform and parameters. The only difference is whether using AC or not.
\cite{le2015mica} uses Kernel descriptor (KDES) for feature extraction  early.
%\cite{reyes2015fine} adopted modified AlexNet to fulfill realistic plant identification task. %\cite{choi2015plant} utilized GoogLeNet and ensemble learning for the recognition.
%\cite{reyes2015fine}   Champ \emph{et al.}    Le \emph{et al.}
the result of \cite{champ2015comparative} is based on a fusion of CNN feature and Fish vector representation using a Bayesian inference framework.
CNN feature and support vector machine (SVM)  classifier are used in \cite{ge2015content}.
\cite{choi2015plant} uses five complementary CNN classifiers  and combines the image classification results with Borda-fuse method. From Table 1, we first observe that the precision of methods using CNN is higher than that of methods using hand-crafted features. It confirms the supremacy of deep learning approaches. State-of-the-art results have been obtained using our proposed approach. The accuracies of "ResNet50+AC" and "Inception v3+AC" are higher than others.  Turning to another comparison, we can see that the results of CNNs using attention cropping augmentation are superior than those of CNNs without attention cropping. Improvement are all demonstrated on the two models. We can see that the improvement is about $4.3\%$ from Table 1. This number indicates that using our proposed novel data augmentation method the performance outperforms the primitive models by a large margin. By conducting attention cropping, "ResNet50+AC" achieves $4.4\%$ improvement in accuracy, compared with original ResNet50. "Inception v3+AC" exceeds original Inception v3 by $4.2\%$. The results show that attention cropping is an efficient data augmentation method. The performance is improved substantially.

\begin{table}[!htbp]
\footnotesize
\caption{Classification results on Oxford Flower}
\label{Table2}
\begin{center}
\begin{tabular}{lc}
\toprule  
Method& Accuracy\\
\midrule  
Angelova and zhu \cite{angelova2013efficient}&0.806\\
Sparse coding\cite{lihua2015two}&0.852\\
MPP \cite{yoo2015multi}&0.913\\
MagNet \cite{rippel2015metric}&0.914\\
Kim\cite{kim2016learning}& 0.945\\
Selective joint fine-tuning\cite{ge2017borrowing}&0.947\\
Ours(ResNet50 without AC)&0.924\\
Ours(Inception v3 without AC)&0.928\\
Ours(ResNet50 with AC)&0.947\\
Ours(Inception v3 with AC)&\textbf{0.951}\\
\bottomrule 
\end{tabular}
\end{center}
\end{table}    

In addition to the specific dataset for real-world plant species recognition,  supplementary experiments on Oxford flower which is for tradition plant species recognition are also provided. Oxford flower consists of 102 different categories of flowers common to the UK. Compared with the dataset for real-world identification,  the background is simple in Oxford flower dataset samples and the objects to be recognized are nearly full of the image. The task of identification is easy and the accuracy can be high even  using common methods. The results are shown in Table 2.

\cite{lihua2015two} proposes a two-layer local constrained sparse coding architecture and achieves a classification performance of $85.2\%$. \cite{yoo2015multi} introduces a multi-scale pyramid pooling and adds a Fisher kernel based pooling layer on top of a pre-trained CNN and obtains $91.3\%$(Acc.). The Acc of \cite{rippel2015metric} is $91.4\%$.
%[magnet] introduce Magnet Loss and the accuracy on Oxford 102 flowers is $91.4\%$.
\cite{ge2017borrowing} introduces selective joint fine-tuning and the accuracy is $94.7\%$. Our results demonstrate new state-of-the-art performance: $95.1\%$ on Oxford 102 flowers. The effectiveness of AC is also observed visually in Table 2. By using AC, the accuracy outperforms that of the original model.  It is increased about $2.3\%$ on Oxford flowers. The boost is $2.2\%$ and $2.3\%$ respectively. This indicates that our scheme is all useful in improving the performance for different types of datasets. It is noted that AC possesses greater advantage for real-world recognition compared with the conventional recognition where the background is simple. The performance is quit significant for the former. One reason is that our scheme of AC makes "hard" samples in real-world recognition "easy"  while those samples in conventional recognition are "easy" originally.

\section{Conclusion}

In this paper, we address real-world species recognition task which is more challenging and makes more sense. Based on deep learning and visual attention, a novel recognition schema is proposed. Images are cropped  in terms of visual attention before recognized. AC helps us focus on our real interested target and remove the interferences. And we apply it as a data augmentation method. It is the first time to crop images in terms of visual attention for data augmentation although there have been many data augmentation methods in deep learning community. An extensive comparative experiments are carried out on different types of datasets including Oxford flower which is a traditional dataset and PlantCLEF which is a specific dataset for real-world identification. Experiments show that new state-of-the-art results have been provided. What is more important, contrastive results indicate that superior improvement is obtained by using AC. And the performance is quit significant especially in realistic identification. What is worth mentioning is that AC can be applied to other recognition tasks and application scenes in the vision community although we mainly focus on real-world plant species recognition in this paper.

In addition, with regard to future work, it would be interesting to investigate a problem that the recognition system has the ability to conduct unknown and never seen categories. And new technologies of machine learning keep future potential role for the interdisciplinary research field of species recognition including real-world species identification for the next few years.
%real-world recognition keeps futrue   .

%, which is necessary for realistic species recognition
% interdisciplinary collaboration   potential latest advances in computer science
%We propose a novel data augmentation method in this paper and name it attention cropping. To the best of our knowledge, it is the first attempt to crop images in terms of visual attention for data augmentation. Considering the complicated background in realistic plants images, multiple saliency detections approach is introduced to generate the region of interesting. Attention cropping makes a significant contribution to data augmentation, which has been a fundamental and important operation for deep learning. We apply attention cropping to real-world plant species identification. Our work can provide a helpful reference for real-world identification.
%
%We validate our method with a series of comparable experiments. Experiments show that the results of methods with attention cropping are superior compared with those methods without attention cropping. Results demonstrate that our method can provide boost on different types of datasets.  Attention cropping is in keeping with our human recognition strategy.     %To the best of our knowledge,
%It is worth noting that although we mainly focus on plant images in this paper, our approach can be obviously applied to other recognition tasks and application scenes in the vision community.
\section*{Acknowledgement}

The work described in this paper is supported by the National Nature Science Foundation of China (NSFC) under Grants  61771346.

\section*{References}

\bibliography{elsarticle-template}

\end{document}